\documentclass[sigconf, nonacm]{acmart}
\AtBeginDocument{%
  }

\newcommand{\bench}{\textsc{CLR-Bench}}
\newcommand{\cmark}{\ensuremath{\checkmark}}
\newcommand{\pmark}{\ensuremath{\circ}}
\newcommand{\histhintavg}{63.5}
\newcommand{\nameoracleavg}{58.0}
\newcommand{\histconvavg}{64.8}

\newcommand{\rgaceincmone}{93.5}
\newcommand{\rgaceincmtwo}{9.5}

\newcommand{\rgdcincmtwo}{14.0}

\newcommand{\rghistincmtwo}{42.0}

\newcommand{\rggepaincmtwo}{4.0}

\newcommand{\stdaceincmtwo}{12.0}

\newcommand{\stddcincmtwo}{11.0}

\newcommand{\stdhistincmtwo}{19.5}

\newcommand{\stdgepaincmtwo}{4.5}

\newcommand{\gepaorigavg}{6.8}
\newcommand{\gepaorigmax}{137}
\newcommand{\gepaneutralselfavg}{49.5}
\newcommand{\gepaneutralhintavg}{66.0}

\usepackage{enumitem}
\usepackage{array}
\usepackage[most]{tcolorbox}
\newtcolorbox{promptbox}[1]{enhanced, breakable, colback=black!2, colframe=black!45, boxrule=0.4pt, arc=1.5pt,
  left=4pt, right=4pt, top=2pt, bottom=2pt, fonttitle=\footnotesize\bfseries, coltitle=black, colbacktitle=black!8,
  title={#1}, fontupper=\footnotesize\ttfamily\raggedright, before skip=4pt, after skip=4pt}
\setlist[itemize]{wide=\parindent, topsep=2pt, partopsep=0pt}

\hypersetup{pdfpublisher={arXiv}}
\begin{document}

\title{More Rules, Longer Lists? Benchmarking Self-Evolving Agents on a Hidden-Rule Curriculum}

\author{Chenjie Qiu}
\affiliation{%
  \institution{Singapore Management University}
  \country{Singapore}}
\email{yausankit@gmail.com}

\author{Boyan Xu}
\affiliation{%
  \institution{Guangdong University of Technology}
  \city{Guangzhou}
  \country{China}}
\email{hpakyim@gmail.com}

\author{Shaobin Shi}
\affiliation{%
  \institution{Guangdong University of Technology}
  \city{Guangzhou}
  \country{China}}
\email{d7inshi@gmail.com}

\author{Bingfeng Chen}
\affiliation{%
  \institution{Guangdong University of Technology}
  \city{Guangzhou}
  \country{China}}
\email{chenbf@gdut.edu.cn}

\author{Ruichu Cai}
\affiliation{%
  \institution{Guangdong University of Technology}
  \city{Guangzhou}
  \country{China}}
\email{cairuichu@gmail.com}

\author{Dongsheng Luo}
\authornote{Corresponding author.}
\affiliation{%
  \institution{Singapore Management University}
  \country{Singapore}}
\email{dsluo@smu.edu.sg}

\renewcommand{\shortauthors}{Qiu et al.}

\begin{abstract}
Self-evolving agents rewrite a persistent state from feedback, but it is largely untested whether this lets them learn
a rule nobody told them. An agent can also pass by writing down each correction as an exception, and its list then
grows with every new rule. We introduce \bench{}, a Curriculum of Latent Rules benchmark. Each chain adds one hidden
rule per level on top of the earlier ones. A rule acts on an opaque label that stands for a named grouping published on
the Web, such as the founding members of an airline alliance. The task gives one example value, practice feedback
reveals one more member and one non-member, and the test checks unseen members and similar non-members. Probes
certify that the base model knows the members of each grouping and does not guess it without feedback, and that the
evidence identifies the grouping once labels are read as named groupings. Four other base models rarely guess the
groupings either. An automated pipeline builds chains from a new data domain, and we release 20 chains over 10
domains, each with four levels that pair a hidden rule with a practice task and a test task. We evaluate six recent
self-evolution methods. With the certifying model they barely learn. The best gain at most 5 points over a frozen
agent and exclude only 6--12\% of the unseen members from the count, fewer than the 20\% reached by simply keeping the
raw practice history. A fixed lesson that states the named-grouping convention lifts the same methods by 14.5 to 16.5
points. Their states hold the revealed member in 75--98\% of cases and almost never the class name, and when practice
exposes a member the list misses, they add that member to the list. With the stronger GPT-6-Sol, ACE and Dynamic
Cheatsheet learn from feedback alone. They reach 78\% accuracy and solve 59--60\% of the single cells, against 53\% and
8\% with the certifying model, and they surpass the raw history. \bench{} therefore measures how well an agent learns,
and this ability differs across models. Code and data are available at
\url{https://github.com/yausankit/clr-bench}.
\end{abstract}


\maketitle

\section{Introduction}
\label{sec:intro}

Large language model agents are increasingly expected to get better with use. Self-evolving methods give an agent a
persistent state that it rewrites after each attempt. The state can be an evolving playbook of
strategies~\cite{zhang2025ace}, an optimized system prompt~\cite{agrawal2025gepa}, a cumulative
cheatsheet~\cite{suzgun2025dynamic}, a library of skills~\cite{evoskill}, a bank of distilled reasoning
memories~\cite{ouyang2025reasoningbank}, or a set of experiences extracted from groups of rollouts without updating
weights~\cite{cai2025trainingfree}. All of these methods expect feedback from the environment to become knowledge the
agent did not have before, and expect that knowledge to carry over to new cases.

Current evaluations rarely test this expectation in isolation, for four reasons. First, a score that rises over a
stream can come from memorizing recurring instances, from following rules the task already states, or from fixing
output formats and tool calls. None of these requires inferring anything new. Second, benchmarks do not measure what
the model knew before the feedback, so a gain is hard to credit to the feedback and a failure hard to blame on the
method. Third, aggregate accuracy hides what was learned: an agent that memorizes the corrected cases can score like
one that infers the rule behind them. Fourth, few benchmarks check that the evidence determines what should be
learned, and a low score may then reflect an ambiguous target. We study a narrower question that is central to
learning from experience. \emph{Given sparse corrective feedback, can a self-evolving agent induce a hidden rule and
apply it to cases it has never seen?}

\begin{figure*}[t]
  \centering
  \includegraphics[width=\textwidth]{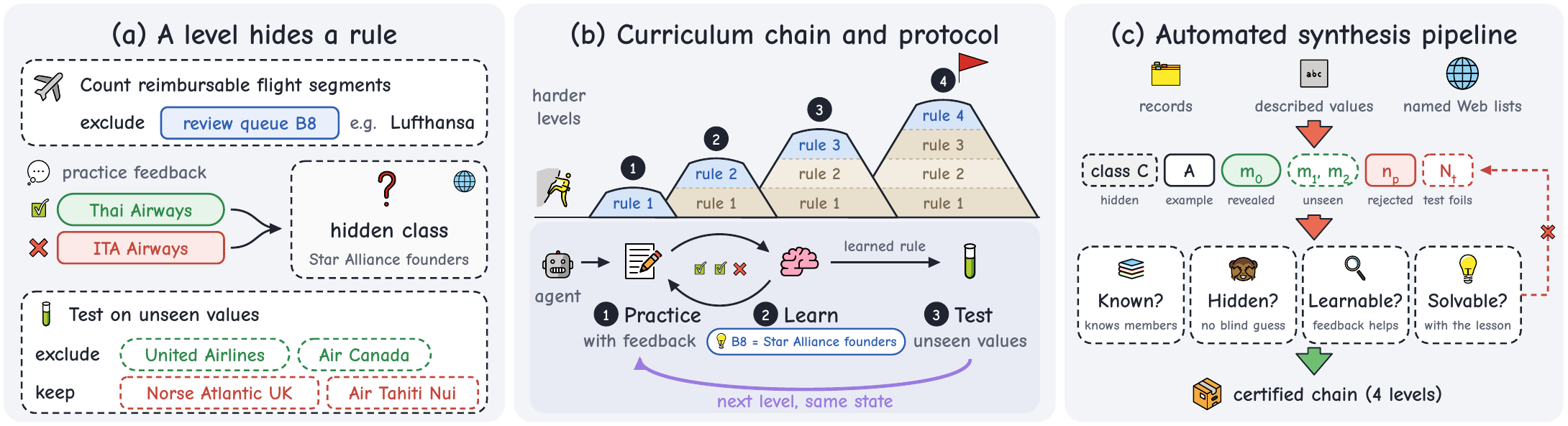}
  \caption{Overview of \bench{}. (a) A level hides a published grouping behind an opaque label, with one example $A$; practice reveals one
  member ($m_0$) and one non-member ($n_p$), and the test checks unseen members ($m_1, m_2$) and test non-members ($N_t$). (b) Rules accumulate over four levels; the agent
  practices, learns, and is tested. (c) Four probes certify each candidate.}
  \label{fig:overview}
\end{figure*}

We introduce \bench{} (\textbf{C}urriculum of \textbf{L}atent \textbf{R}ules Benchmark) to answer it. Its chains
consist of aggregation tasks over realistic records, such as counting reimbursable flight segments per department or
reviewable medication records per patient. Each level adds a rule that excludes records whose field value falls under
an opaque label, for instance ``segments in review queue B8 are not counted.'' The task gives one example value for
the label, \emph{Lufthansa}, and short factual descriptions of the values that appear. The label stands for a
real-world named grouping taken from public sources, here the founding members of Star Alliance, but the task never
names it. Practice feedback comes from small edits of a public example. The agent learns that \emph{Thai Airways}
belongs to review queue B8 like \emph{Lufthansa} and that \emph{ITA Airways} does not. The test then asks it to apply
the rule to the remaining members, such as \emph{United Airlines}, which never appeared in feedback, and to keep
counting similar-looking non-members. To succeed, the agent has to turn two pieces of evidence into the right category
and write that category into its state.

The hidden classes come from the Web. Agents increasingly act on Web data and services, and the categories behind
their tasks are groupings that organizations and standards bodies publish and maintain online: the airlines of an
alliance, the countries of an economic bloc, the weaknesses of a security taxonomy, the companies of a stock index.
A task rarely spells such a grouping out, and an agent meets it through examples and corrections. Recovering a Web
grouping from a few seed members is the classic problem of set expansion~\cite{wang2007seal,zhang2020cgexpan}, and
learning a concept from a few positive examples has long been studied in cognitive
science~\cite{tenenbaum1999bayesian}. \bench{} moves this problem into the self-evolution loop. The seeds arrive as
feedback, and the agent should store the grouping so that later decisions use it. Hidden classes are published
groupings or values tied to their members, for instance the currencies of BRICS member states. The benchmark's lists
define them and record their sources and snapshot dates.

Practice and test reward different things by design. An agent can pass practice by recording the two revealed values
as exceptions. Only an agent that asks which category explains a correction also handles the unseen members. Earlier
rules stay in force, so the gap grows with depth. A list-keeping agent adds another list at every level and misses
the unseen members of every rule, while an agent that writes categories carries them forward. Real feedback has the same shape: an agent is corrected on
individual cases and has to infer the rule behind them.

Three design choices make the benchmark diagnostic. The evidence is sufficient under one convention. If labels denote
named groupings, the example, the revealed member, and the rejected non-member single out one grouping of the
domain's library, and the base model knows which of the described values belong to it. The evidence is also necessary.
Language-model probes certify during construction that the label and the example alone do not give the class away.
Finally, the same tasks run under several conditions that show where learning fails. A \emph{frozen} agent never
updates. Each method learns from feedback alone (\emph{self}), and again with a fixed \emph{lesson} in its update step
that states the convention and a procedure for using the evidence without naming any class. A method-free reference
keeps the raw practice history in the prompt. Value-level diagnostics tell under-generalization, where unseen members
stay in the count, apart from over-generalization, where non-members are removed.

An automated pipeline assembles and certifies chains from a new data domain. A chain holds the records of one domain,
with entities such as departments, employees, claims, and flight segments, a base task that counts records per
entity, and four levels. Each level adds a rule over one field with an opaque label, describes every value involved,
and comes with a practice task and a test task. We release 20 chains, two in each of 10 domains that range from travel
expense audits and pharmacy reviews to system logs, chemical imports, and port state control. Together they contain 80 hidden rules, 80 practice tasks,
and 80 test tasks.

Our contributions are as follows.

(1) \textbf{Task definition.} We define curriculum chains in which each level hides a rule over an opaque label that
stands for a published grouping. Practice and test ask for different outputs, so only an inferred rule transfers
(Section~\ref{sec:task}).

(2) \textbf{Probe-certified construction.} An automated pipeline builds chains from a new data domain and uses
language-model probes to certify that the model knows the members of each class and does not guess the class without feedback. Under the
named-grouping convention the evidence identifies the class. We release 20 chains over 10 domains with 80 hidden rules,
each paired with a practice task and a test task (Section~\ref{sec:construction}).

(3) \textbf{Diagnostic metrics and conditions.} Single-cell accuracy and value-level diagnostics separate under- from
over-generalization. A frozen agent, the raw history, and a lesson that states the convention bound what the evidence
supports (Section~\ref{sec:metrics}).

(4) \textbf{Evaluation of six methods with two base models.} With the certifying model, ACE, GEPA, and Dynamic
Cheatsheet gain 2 to 5 points and exclude fewer unseen members from the count than the raw history does. Their states
record corrections as a growing list of exceptions. A fixed lesson in the update step lifts them by 14.5 to 16.5
points, and its convention part alone lifts ACE and Dynamic Cheatsheet by 13 to 19 points, or by 11 points when it is
stated at level 1 only. With the stronger GPT-6-Sol, ACE and Dynamic Cheatsheet solve 59--60\% of the single cells from
feedback alone, against 8\% with GPT-6-Luna. The benchmark therefore separates learning ability across models
(Sections~\ref{sec:experiments} and~\ref{sec:analysis}).

\section{Related Work}
\label{sec:related}

\begin{table*}[t]
  \begin{minipage}[t]{0.485\textwidth}
  \caption{Comparison with related benchmarks.}
  \label{tab:compare}
  \small
  \setlength{\tabcolsep}{2.2pt}
  \begin{tabular*}{\linewidth}{@{\extracolsep{\fill}}llccccc@{}}
    \toprule
    Benchmark & Venue & Fb & State & Hidden & Prior & Diag \\
    \midrule
    LLF-Bench~\cite{cheng2023llfbench} & arXiv'23 & \cmark & -- & -- & -- & -- \\
    MINT~\cite{wang2024mint} & ICLR'24 & \cmark & -- & -- & -- & -- \\
    StreamBench~\cite{wu2024streambench} & NeurIPS'24 & \cmark & \cmark & -- & -- & -- \\
    Evo-Memory~\cite{wei2025evomemory} & arXiv'25 & \cmark & \cmark & -- & -- & -- \\
    EvoMemBench~\cite{wang2026evomembench} & arXiv'26 & \cmark & \cmark & -- & -- & -- \\
    EvoAgentBench~\cite{gao2026evoagentbench} & arXiv'26 & \cmark & \cmark & -- & -- & -- \\
    MemoryAgentBench~\cite{hu2025memoryagentbench} & ICLR'26 & -- & \cmark & \pmark & -- & -- \\
    WILT~\cite{banatt2024wilt} & NeurIPS-W'24 & \cmark & -- & \cmark & \pmark & -- \\
    InductionBench~\cite{hua2025inductionbench} & ACL'25 & -- & -- & \cmark & -- & -- \\
    \midrule
    \bench{} (ours) & -- & \cmark & \cmark & \cmark & \cmark & \cmark \\
    \bottomrule
  \end{tabular*}
  \\[2pt]{\footnotesize Fb: learns from feedback on its own attempts; State: knowledge persists across tasks; Hidden: the rule is never stated; Prior: prior knowledge is certified; Diag: value-level diagnostics. \cmark{} yes, \pmark{} partly, -- no.\par}
  \end{minipage}\hfill
  \begin{minipage}[t]{0.485\textwidth}
  \caption{Evaluated self-evolution methods.}
  \label{tab:methods}
  \small
  \setlength{\tabcolsep}{2.6pt}\renewcommand{\arraystretch}{1.55}
  \begin{tabular*}{\linewidth}{@{\extracolsep{\fill}}llccccc@{}}
    \toprule
    Method & Venue & Reflect & Incr. & Gate & Retr. & Code \\
    \midrule
    ACE~\cite{zhang2025ace} & ICLR'26 & \cmark & \cmark & -- & -- & -- \\
    GEPA~\cite{agrawal2025gepa} & ICLR'26 & \cmark & -- & \cmark & -- & -- \\
    Dynamic Cheatsheet~\cite{suzgun2025dynamic} & EACL'26 & \cmark & -- & -- & -- & -- \\
    Training-Free GRPO~\cite{cai2025trainingfree} & arXiv'25 & \cmark & \cmark & \cmark & -- & -- \\
    EvoSkill~\cite{evoskill} & arXiv'26 & \cmark & \cmark & \cmark & \pmark & \cmark \\
    ReasoningBank~\cite{ouyang2025reasoningbank} & ICLR'26 & \cmark & \cmark & -- & \cmark & -- \\
    \bottomrule
  \end{tabular*}
  \\[2pt]{\footnotesize Reflect: an LLM reflects on failures to write the update; Incr.: the state is edited, not rewritten; Gate: an update is kept only if a condition holds; Retr.: the solver sees only retrieved entries; Code: the state holds executable code.\par}
  \end{minipage}
\end{table*}

\paragraph{Self-evolving agents.}
Several methods let a frozen language model improve across episodes by rewriting a persistent, non-parametric state.
Reflexion keeps verbal self-reflections in an episodic buffer~\cite{shinn2023reflexion}. ExpeL extracts insights from
successful and failed trajectories~\cite{zhao2023expel}. Voyager and EvoSkill grow libraries of executable
skills~\cite{wang2023voyager,evoskill}, and Agent Workflow Memory induces reusable workflows from past
episodes~\cite{wang2024awm}. More recent methods change what is stored and how it is updated. Dynamic Cheatsheet
rewrites a cumulative cheatsheet after every attempt~\cite{suzgun2025dynamic}, and ACE curates an itemized playbook with
incremental edits~\cite{zhang2025ace}. ReasoningBank distills strategies from both successes and
failures~\cite{ouyang2025reasoningbank}. Training-Free GRPO extracts experiences by contrasting rollouts within a
group~\cite{cai2025trainingfree}, and GEPA evolves the system prompt by reflective search over a Pareto
front~\cite{agrawal2025gepa}. We evaluate six of these methods, and Table~\ref{tab:methods} summarizes how each one
writes and uses its state. They all assume that feedback contains knowledge worth keeping. \bench{} tests whether they
extract it when the knowledge is a category that the feedback only points at.

\paragraph{Benchmarks for learning from feedback.}
Self-evolving methods are usually evaluated on coding, web navigation, app-based agent tasks, and financial
reasoning~\cite{zhang2025ace,ouyang2025reasoningbank,wang2024awm}. StreamBench treats the setting as online learning
from a feedback stream~\cite{wu2024streambench}, and LLF-Bench and MINT test learning from language
feedback~\cite{cheng2023llfbench,wang2024mint}. Recent benchmarks for self-evolving agents measure memory over task
streams~\cite{wei2025evomemory,wang2026evomembench,hu2025memoryagentbench} and the transfer of learned procedures across
tasks~\cite{gao2026evoagentbench}. These benchmarks ask whether scores rise over a stream. \bench{} asks what was
learned and tests it on values that never appear in feedback. Table~\ref{tab:compare} compares them with \bench{}.

\paragraph{Inductive reasoning and rule learning.}
Rule induction from few examples has long served to evaluate intelligence, from concept learning to the Abstraction
and Reasoning Corpus~\cite{chollet2019measure}. Qiu et al.\ find that language models propose good hypotheses but
struggle to apply them~\cite{qiu2023phenomenal}. InductionBench reports failures on the simplest classes of string
functions~\cite{hua2025inductionbench}. WILT asks models to discover a hidden rule by proposing test cases over several
turns~\cite{banatt2024wilt}, and DiscoveryWorld hides regularities in a virtual environment for discovery
agents~\cite{jansen2024discoveryworld}. Two methods come closest to what \bench{} asks of an agent. Hypotheses-to-Theories
builds a rule library on training problems and reuses it at test time~\cite{zhu2023htt}, and hypothesis search proposes
natural-language hypotheses and checks them as programs~\cite{wang2024hypothesis}. Both learn from labeled examples
given in a batch. In \bench{}, evidence reaches an agent's update step one correction at a time, and we evaluate methods
built for that loop. The feedback consists of counterexamples to the agent's program, as in counterexample-guided
synthesis~\cite{solarlezama2006sketching}, and the groupings consistent with the evidence form a version
space~\cite{mitchell1982generalization}. Most of the benchmarks above evaluate one episode of reasoning or exploration.
\bench{} places induction inside an agent's interaction loop. The evidence is feedback on the agent's own programs, and
the rule has to be stored and then applied to a separate test task with a different output. Because the hidden classes
are real groupings that the model may already know, success depends on linking feedback to world knowledge.

\paragraph{Set expansion and concept learning.}
Expanding a few seed entities into the full Web grouping they belong to is a classic Web mining task. It has been
solved with wrapper induction over Web pages~\cite{wang2007seal} and with class names generated by language
models~\cite{zhang2020cgexpan}. Cognitive science models concept learning from a few positive examples as Bayesian
inference over nested hypotheses, in which the size principle favors the smallest consistent
concept~\cite{tenenbaum1999bayesian}. \bench{} gives this problem to an agent that receives the seeds one at a time as
feedback and keeps the inferred grouping in its state. Its diagnostics measure the two errors these fields study:
expanding too little, which misses unseen members, and expanding too much, which includes non-members.

\section{Preliminaries and Task Definition}
\label{sec:task}

\subsection{Curriculum Chains}
A chain is set in a data domain with entities and records, such as departments, employees, claims, and flight
segments. Its base task asks for a program that counts records per entity. A chain has four levels, and the rules
accumulate. Level $j$ adds a rule over a field $f_j$, its \emph{trap field}, with an opaque label $\ell_j$, for example
``segments in review queue B8 are not counted.'' We call a position in a chain a \emph{slot}, and \bench{} has 80
slots. Behind each label is a hidden class $C_j$, a named grouping published on the Web
(Section~\ref{sec:construction}). The task never names $C_j$. It gives one example $a_j \in C_j$ and a one-sentence
factual description of every value that can appear in the level's inputs, including the unseen members and the
non-members, so an agent that knows the class can recognize its members. Each class also has three further members
$m_0, m_1, m_2$, a practice non-member $n_p$, and seven test non-members $N_t$. The non-members come preferentially
from competing groupings that also contain the example (Figure~\ref{fig:overview}a).

\subsection{Practice and Test}
At level $k$ the agent practices on the same records and rules, but the practice task asks for the largest entity
total. The agent may submit up to three programs. Each program runs on a public example and on $2k+1$ hidden inputs.
One hidden input moves the example values to other records. For each rule $j \le k$, one input inserts $m_0$ and
another inserts both $m_0$ and $n_p$. Every failing input comes back as a counterexample that says, for each changed
value, whether it belongs to the label like the example. The first attempt usually fails, because $m_0$ has not been
shown yet. Appendix~\ref{sec:example} reproduces a task and its feedback. The method runs its update step after every
attempt, and a gated method may reject the update. Practice at a level ends at the first passing attempt or after
three attempts. The state at that point writes the test program, which returns the sum of the entity totals.

The test has five hidden inputs, which we call \emph{cells}. Each cell is a full set of records. The $5-k$ \emph{clean
cells} contain only example values. Each of the $k$ \emph{single cells} adds, for one rule, its two unseen members and
its seven test non-members. A program that applies every rule only to its example scores $(5-k)/5$, which gives the
frozen staircase of 0.80, 0.60, 0.40, and 0.20. Any gain above it comes from applying a rule to unseen members while
keeping the non-members. Recording the example and $m_0$ is enough to pass practice, so practice never rewards
generalization directly. The test measures whether the agent infers which category explains its corrections, which is
what a deployed agent would have to do.

\section{Benchmark Construction}
\label{sec:construction}

\begin{figure*}[t]
  \centering
  \includegraphics[width=\textwidth]{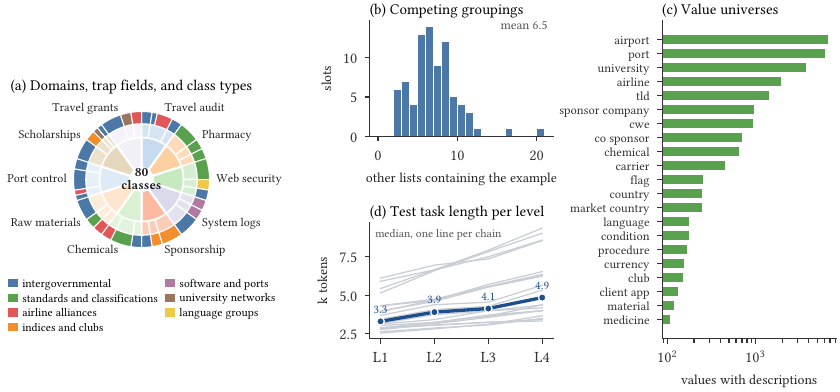}
  \caption{Composition of \bench{}. (a) Hidden classes by domain (inner), trap field (middle), and kind of grouping
  (outer). (b) Other published lists that contain each example. (c) Size of each value universe. (d) Length of the test
  task per level.}
  \label{fig:composition}
\end{figure*}

Writing hidden-rule chains by hand does not scale, and it invites shortcuts such as classes that the example gives
away or members that share a visible prefix. We therefore build chains with a pipeline that turns a domain
description into certified chains (Figure~\ref{fig:overview}c). The domain is the only manual input.

\subsection{Domain Inputs and Candidates}
A domain provides three inputs. The records come from a seeded generator or a public research release; two domains
use Synthea's synthetic patient records and Loghub system logs, and no personal data is involved. Each trap field has
a universe of values with one-sentence descriptions. A library of named lists supplies the candidate classes, such as
organization rosters, alliance memberships, index constituents, official classifications, and sections of standards,
each with its source and snapshot date (Table~\ref{tab:domains}). For each level the pipeline draws a class $C$ and
picks the example, three members, the practice non-member, and seven test non-members. Several constraints block
shortcuts. The example must lie in at least two other lists, and no single attribute may separate members from
non-members. Different levels use disjoint value pools, and non-members come preferentially from competing lists that
contain the example. Disputed or recently changed members are never used.

\begin{table}[t]
  \caption{Probes that certify every hidden class (sampled over the three choices of the revealed member).}
  \label{tab:probes}
  \small
  \setlength{\tabcolsep}{3pt}
  \begin{tabular}{@{}lp{0.47\columnwidth}l@{}}
    \toprule
    Probe & Model is given & Requirement \\
    \midrule
    Known & pool values, one feedback, lesson & names it, soft score $\ge$0.8 \\
    Hidden & label and example, no feedback & does not recover the class \\
    Learnable & practice feedback, no lesson & assigns 1/6--2/3 of unseen \\
    Solvable & feedback and the lesson & soft score $\ge$0.8 \\
    \bottomrule
  \end{tabular}
\end{table}

\subsection{Probe Certification}
Each candidate passes four probes run with the evaluated model (Table~\ref{tab:probes}). The Known probe runs first,
on every candidate list. It shows the model the list's pool values with their descriptions, one round of feedback,
and the lesson, and asks it to name the grouping and mark which values belong to it. The soft score is the share of
pool values whose membership the model gets right, and the probe requires at least 0.8 on average and 0.6 in every
rotation of the revealed member. A list passes only if the model both identifies the grouping and knows its members.
The other three probes run in the context of the full chain. The Hidden probe gives only the label and the example,
and the model must not recover the class. The Learnable probe gives practice feedback alone and requires the model to
assign between 1/6 and 2/3 of the unseen members. The Solvable probe adds the lesson and requires a soft score of at
least 0.8 on the remaining pool values. A candidate that fails is replaced. Together the probes
place each class where feedback matters, between classes the model guesses outright and classes it cannot learn.
Built chains are checked for consistency and reproduce bit for bit from their seeds. A frozen agent scores 0.80, 0.57,
0.34, and 0.20 on them, close to the designed staircase.

\subsection{Benchmark Statistics}
Figure~\ref{fig:composition} shows the composition of \bench{}. The 20 chains cover ten domains with two chains
each. Every domain contributes three trap fields, whose value universes range from about 100 values for medicines and
materials to several thousand for airports, ports, and universities. The 80 hidden classes span seven kinds of
published grouping. Intergovernmental memberships make up the largest share with 35 classes, followed by 18 standards
and official classifications, 10 airline alliances, and 8 stock indices and sports clubs. The rest are software
products and ports, university networks, and language families. Every example also belongs to other published lists,
6.5 on average and up to 20, so the evidence has to separate the class from real competitors. Under the named-grouping
convention it does. Tasks grow with depth as rules and records accumulate, from a median of about 3.3k tokens at level
1 to 4.9k at level 4. Table~\ref{tab:classes} lists all classes.

\section{Evaluation Metrics and Conditions}
\label{sec:metrics}

\subsection{Level and Single-Cell Accuracy}
The level score is the share of the five hidden test inputs for which a program returns the correct sum. The share of
clean cells falls with depth, so we also report single-cell accuracy, the share of correct single cells pooled over the
20 chains and levels 1--4. It is zero for an agent that applies every rule only to the example, and the mixture of
cells does not affect it.

\subsection{Value-Level Diagnostics}
For every value in a class pool we build an input that adds only that value to the example values. We run the level-$k$
test program on it for every rule $j \le k$ and record two outcomes. For a member we record whether it is excluded,
written $\mathrm{inc}$, or $\mathrm{inc}_u$ for the two unseen members $m_1$ and $m_2$. For a non-member we record
whether it is kept, written $\mathrm{exc}$. Both are pooled over chains, levels, and rules. A low $\mathrm{inc}_u$
signals under-generalization, and a low exc signals over-generalization.

\subsection{Conditions}
The main comparison varies two things, the method and the base model (Table~\ref{tab:conditions} in the appendix). Each
method runs \emph{self}, learning from feedback alone, and again with a \emph{lesson} added to its update step. The
lesson is a fixed text that names no class. It states the benchmark's convention, namely that a label denotes a
well-known named grouping that contains the example and the confirmed member and excludes the rejected value, and it
gives a procedure for applying this convention. Certification requires the lesson to make every class learnable, so it
marks what the evidence can support. A method-free \emph{history} reference puts the raw practice programs and feedback
into the solver prompt, with or without the lesson. Under \emph{rewarded practice}, the hidden practice inputs of the
second and third attempts also insert $m_1$. An exception list built from the first attempt's feedback then fails a
second time, and only $m_2$ stays unseen. The appendix reports two further reference conditions, the parts of the
lesson and the class names.

\section{Experimental Setup}
\label{sec:experiments}

\subsection{Baselines}

\subsubsection{Self-evolution methods}
We evaluate six self-evolution methods that differ in what they store and how they update it
(Table~\ref{tab:methods}). We follow the official prompts and algorithms and replace only the task-specific
placeholders.

\subsubsection{Reference conditions}
A \emph{frozen} agent writes every test program with an empty state and sets the floor. The method-free \emph{history}
reference puts the raw practice programs and feedback into the solver prompt. It shows what the evidence supports
without any update step. The \emph{lesson} in the update step shows what the evidence supports under the benchmark's
convention (Section~\ref{sec:metrics}).

\subsection{Experimental Settings}
All methods, the frozen agent, and the references use GPT-6-Luna with high reasoning effort.
Section~\ref{sec:basemodel} reruns the frozen agent, the history reference, and all six methods with GPT-6-Sol at high
effort, both from feedback alone and with the lesson. At each level a method gets up to three practice attempts and
updates its state after every attempt. An attempt is one submitted program, and Training-Free GRPO samples its group of
four rollouts within an attempt. The state reached after practicing level $k$ writes the level-$k$ test program, which
is scored once. The state carries over from one level to the next, and the history reference sees every earlier
practice program and its feedback from all levels so far. Practice feedback is explicit. For every value that a hidden
practice input changed, it says whether the value belongs to the label, for example ``X belongs to review queue B8,
like Lufthansa; its records are not counted.'' Each condition runs once per chain, and we report averages over the 20
chains. Besides the level score we report single-cell accuracy and the diagnostics $\mathrm{inc}_u$ and exc
(Section~\ref{sec:metrics}).

\section{Experimental Results}
\label{sec:analysis}

\subsection{Main Results}
\subsubsection{Overall results}

\begin{figure*}[t]
  \centering
  \includegraphics[width=\textwidth]{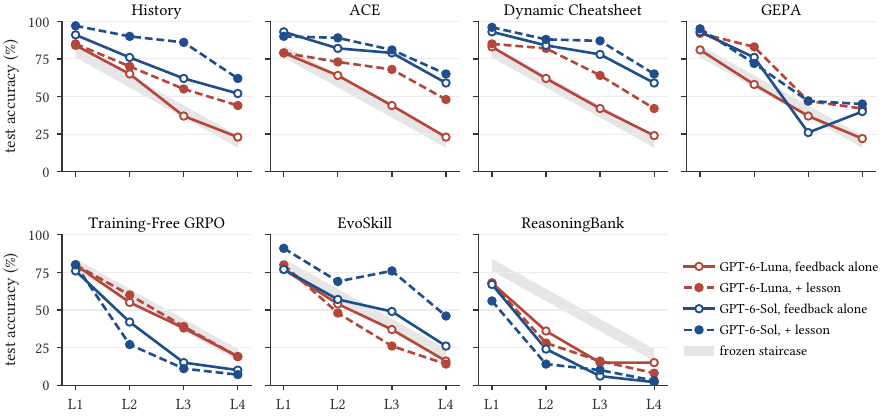}
  \caption{Test accuracy per level (20 chains) from feedback alone (open) and with the lesson (filled), with GPT-6-Luna
  (red) and GPT-6-Sol (blue). Gray band: the designed staircase (80, 60,
  40, 20) of an agent that applies each rule only to the example.}
  \label{fig:main}
\end{figure*}

Figure~\ref{fig:main} shows the accuracy of every method per level with both base models, and
Table~\ref{tab:mainfull} in the appendix lists every condition per level.

With GPT-6-Luna and feedback alone, the methods barely learn, and none of them generalizes better than the raw
evidence. The frozen agent passes 95.5\% of the clean cells and none of the single cells, so any point above it comes
from applying a rule beyond the example. ACE, GEPA, and Dynamic Cheatsheet gain only 1.8 to 5.0 points. They solve
3--8\% of the single cells, exclude 6--12\% of the unseen members, and stay on the staircase at every level. The
method-free history reference keeps every practice program and its feedback in the prompt. Its accuracy is about the
same at 52.3\%, and it excludes more unseen members, 20\%. The update steps make programs more precise, keeping 92--95\%
of the non-members where the history keeps 82\%, but they add no generalization.

Three methods fail because of the way they accept updates. Training-Free GRPO extracts experiences only from groups
with mixed outcomes, and EvoSkill accepts a skill only if it raises the practice score. Practice outcomes are
all-or-nothing. As a result, no Training-Free GRPO state ever contains an experience, and EvoSkill accepted no skill on
9 of the 20 chains; a third of its 449 proposals were malformed. Every Training-Free GRPO state is empty with both base
models, and 70\% of EvoSkill's states hold no skill with GPT-6-Luna, against 14\% with GPT-6-Sol. ACE and Dynamic
Cheatsheet, by contrast, update after every attempt. ReasoningBank retrieves generic strategies such as ``infer the
full category behind a field-note example,'' and the solver then excludes values on the strength of their descriptions
alone. It keeps only 51\% of the non-members and falls 14 points below the frozen agent.

\subsubsection{Effect of the base model}
\label{sec:basemodel}
We rerun the frozen agent, the history reference, and all six methods with GPT-6-Sol, from feedback alone and with
the lesson (Figure~\ref{fig:main}, blue, and Table~\ref{tab:sol}). Without feedback, Sol guesses a class in only 1 of
the 80 level scores. Its frozen agent even falls below the staircase to 32.5\%, because its programs also drop
non-members that fit a guessed attribute category such as ``airlines based in European countries.''

From feedback alone, ACE and Dynamic Cheatsheet with Sol reach 78.3\% and 78.5\%. They solve 59--60\% of the single
cells and exclude 68--76\% of the unseen members, where Luna managed 8\% and 12\%. They also overtake the history
reference, which reaches 70.3\% and 49\% of the single cells. Most of the margin comes from keeping more non-members,
90--95\% against 83\%. Dynamic Cheatsheet also excludes more unseen members than the history, 76\% against 68\%, while
ACE excludes as many. The lesson adds only 3.0 and 5.5 points with Sol, compared with 14.5 and 15.5 with Luna. The gap
between the two models widens with depth, and at level 4 the self condition scores 59\% with Sol and 24\% with Luna.
The exception lists seen with Luna are what an update step writes when the model does not propose a category. A model
that proposes the category writes the category instead. Difficulty was calibrated on the certifying model. Sol scores
above 80\% on the first two levels, and level 4 still leaves 41 points to gain.

\begin{figure}[t]
  \centering
  \includegraphics[width=\columnwidth]{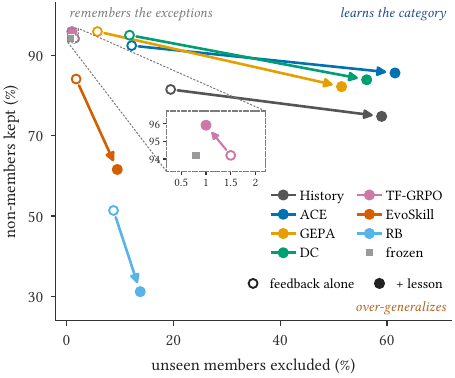}
  \caption{Generalization against specificity (GPT-6-Luna, levels 1--4 pooled): each method from feedback alone (open)
  to + lesson (filled). The inset enlarges the top-left corner,
  where Training-Free GRPO and the frozen agent overlap.}
  \label{fig:plane}
\end{figure}

\begin{figure}[t]
  \centering
  \includegraphics[width=\columnwidth]{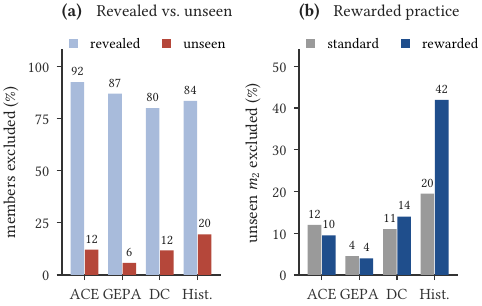}
  \caption{GPT-6-Luna, learning from feedback alone. (a) Exclusion of the revealed member $m_0$ and of the unseen members
  $m_1, m_2$. (b) Exclusion of $m_2$, which is never shown, under standard and rewarded practice (where $m_1$ is revealed
  from the second attempt).}
  \label{fig:results}
\end{figure}

A stronger model does not help every update mechanism. GEPA and EvoSkill with Sol exclude more unseen members, 48\% and
41\%, but keep only 66\% and 53\% of the non-members and end at 58.8\% and 52.3\%. Training-Free GRPO still extracts no
experience, and ReasoningBank's strategies drop even more non-members, keeping 21\%. Both end below their Luna scores,
at 35.8\% and 24.8\%. A model that proposes categories helps the methods whose update step records them and makes the
others over-generalize. With the lesson, Sol raises History to 83.8\%, GEPA to 64.8\%, and EvoSkill to 70.5\%.
Training-Free GRPO and ReasoningBank stay below their feedback-only scores at 31.2\% and 20.8\%. When the update step
does not store a rule, neither a better model nor the convention helps.

Other models cannot guess the hidden classes either (Table~\ref{tab:backbones} in the appendix). Without feedback,
frozen agents with four further base models stay at or below the staircase, and at most 4 of their 80 level scores lie
above it. Scores below the staircase come from programs that also drop values the clean cells must count. When told
the class names, these models apply them about as imprecisely as GPT-6-Luna, solving 40--45\% of the single cells and
keeping 51--63\% of the non-members. The gap between knowing a grouping and applying it is therefore not specific to
one model. Table~\ref{tab:sol} in the appendix lists every learning condition for both base models.

\subsection{Analysis}

\begin{figure*}[t]
  \centering
  \includegraphics[width=\textwidth]{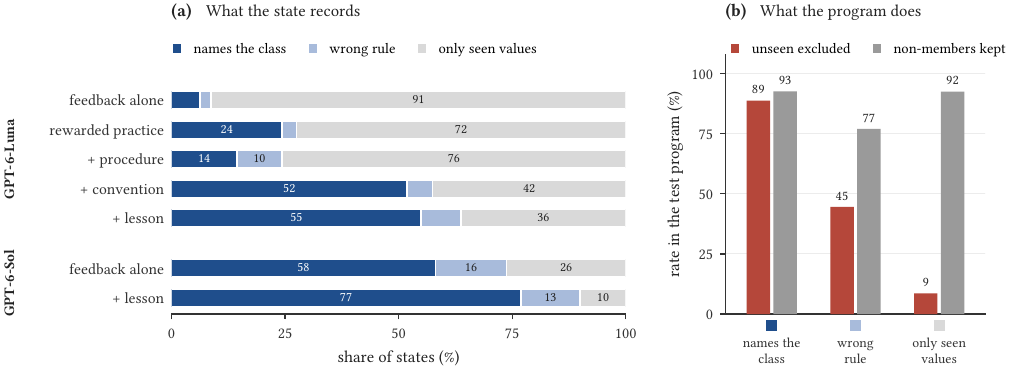}
  \caption{(a) What the state says about each level's label (ACE and Dynamic Cheatsheet, 160 states per row, annotated
  without test results; names the class: the class, a broader grouping, or its members listed; wrong rule: another
  grouping or an attribute rule). (b) Unseen members excluded and non-members kept by the program, by state content,
  over all annotated states.}
  \label{fig:states}
\end{figure*}

\subsubsection{Why do the methods not learn?}
With the lesson in their update step, ACE, GEPA, and Dynamic Cheatsheet reach 67.0\%, 66.0\%, and 68.3\% and solve
39--46\% of the single cells, on par with History + lesson at \histhintavg\%. The same evidence therefore suffices to
learn the rules. The gap to the self condition grows as rules accumulate. At level 4, ACE scores 48\% with the lesson
and 23\% without it, GEPA 42\% and 22\%, and Dynamic Cheatsheet 42\% and 24\%. The lesson helps only methods whose
update step stores a rule. Training-Free GRPO stays at 49.5\% because its gate never opens. EvoSkill and ReasoningBank
lose accuracy with the lesson, falling to 42.0\% and 30.0\% and keeping only 62\% and 31\% of the non-members. Once told
that labels denote named groupings, they remove whole groupings they guess. The gains hold across domains
(Table~\ref{tab:domainres} in the appendix). The lesson raises ACE in 8 of the 10 domains and Dynamic Cheatsheet in 9,
and web security is the exception for both.

In the diagnostics, ACE excludes the member revealed in practice 92.5\% of the time but the two unseen members only
12.2\% of the time. Dynamic Cheatsheet and GEPA behave the same way, at 80.0\% against 11.8\% and 87.0\% against 5.8\%
(Figure~\ref{fig:results}a). The states record the revealed member as an exception. Figure~\ref{fig:plane} shows that
all methods learning from feedback alone sit in the corner of high specificity and low generalization. The lesson moves
ACE, GEPA, and Dynamic Cheatsheet toward the learned category and pushes EvoSkill and ReasoningBank toward
over-generalization. By string match, the revealed member appears in 75--98\% of the final states of ACE, GEPA, and
Dynamic Cheatsheet, and the full class name in at most 3\%, or 8--18\% with the lesson. In a chain whose class is the
founding members of Star Alliance, the ACE playbook after level 1 reads: ``the confirmed members include Lufthansa and
Thai Airways; ITA Airways is confirmed not to be a member and remains countable. Apply membership by the exact airline
value, not by name similarity.'' With the lesson it reads: ``B8 is Star Alliance: exclude Lufthansa and Thai Airways,
but count ITA Airways.'' The lesson state names Star Alliance, a slightly broader grouping than its founding members,
which fits the lower exc of the lesson condition. Test non-members taken from competing groupings are kept as often as the others, 92.2\% against
92.8\% for ACE. When asked directly in a single prompt, the bare model assigns 44\% of the unseen members from the same
feedback, as the Learnable probe shows.

We annotated the state written after each level by what it says about that level's label. The annotator was
GPT-6-Luna, which saw the hidden class and the practice evidence but no test result
(Appendix~\ref{sec:annotation}). With GPT-6-Luna and feedback alone, 91\% of the ACE and Dynamic Cheatsheet states hold
only the values seen in practice or nothing about the label, and 6\% name the class or list its members. With
GPT-6-Sol, 58\% name the class or list its members (Figure~\ref{fig:states}a). State content predicts what the program
does. When a state names the class or lists its members, the program excludes 89\% of the unseen members and keeps 93\%
of the non-members. When the state holds only the seen values, the program excludes 9\% (Figure~\ref{fig:states}b). The
methods fail to turn corrections into a category, and they apply a category well once it is stored.

To see which part of the lesson matters, we split it inside the update step and give both parts under the same
neutral heading, ``UPDATE NOTE'' (Table~\ref{tab:variants}). The procedure alone tells the method to decide the class
behind the example and apply it to every described value. It leaves ACE and Dynamic Cheatsheet at 55.0\%, a gain of
2.5 and 2.2 points. The convention alone says that a label denotes a well-known named grouping chosen to fit the
corrections. It lifts them to 71.5\% and 65.8\%, a gain of 19.0 and 13.0 points, so its effect does not depend on the
authority of the original heading. With the raw history, the convention alone gives \histconvavg\%, as much as the whole
lesson at \histhintavg\%, while the procedure alone adds 2.0 points. Giving the class names in the update step together
with the feedback lifts ACE and Dynamic Cheatsheet to 73.5\% and 75.8\% and to 52--57\% of the single cells. A solver
told the names without any feedback reaches only \nameoracleavg\%, with 40\% of the single cells and 58\% of the
non-members kept. The methods fail at identifying the category, and once it is identified, the feedback sets its
boundary. The state annotations show the same split. With the procedure alone, 76\% of the states hold only the seen
values or nothing, and with the convention, 52\% name the class or list its members. Dynamic Cheatsheet with the whole
lesson under the neutral heading stays at 53.8\%. Its states show why: 69\% of them no longer mention the label,
because the cheatsheet is rewritten as a whole and the rule gets dropped.

To test whether the methods learn a reusable inference policy, we give the convention in the update step at level 1
only and remove it from level 2 on (Table~\ref{tab:variants}). ACE and Dynamic Cheatsheet reach 63.8\% and 64.3\%,
between learning from feedback alone at 52.5\% and 52.8\% and receiving the convention at every level at 71.5\% and
65.8\%. At levels 2--4, whose classes never came with the convention, ACE scores 75, 58, and 39\%, compared with 64, 44,
and 23\% from feedback alone. Most of the effect of one statement persists in the state. Where it fades, the methods
fall back on attribute rules, for example ``airports in Africa'' for airports in OPEC member states.

\subsubsection{Does rewarded practice change what the methods learn?}
An exception list passes standard practice, and declining to generalize from two positives and one negative could be
a cautious choice. Under rewarded practice a further member, $m_1$, appears from the second attempt on. ACE, GEPA, and
Dynamic Cheatsheet then exclude the newly revealed $m_1$, ACE in \rgaceincmone\% of cases, but they still miss the
member they never saw. Their inc on $m_2$ is \rgaceincmtwo\%, \rggepaincmtwo\%, and \rgdcincmtwo\%, close to the
\stdaceincmtwo\%, \stdgepaincmtwo\%, and \stddcincmtwo\% of standard practice (Figure~\ref{fig:results}b). With the same
signal, the history reference more than doubles its inc on $m_2$, from \stdhistincmtwo\% to \rghistincmtwo\%. The raw
history uses the evidence that a list fails, and the methods do not.

\subsubsection{Failure modes, retention, and cost}
In the self condition, 83--97\% of the single cells of ACE, GEPA, Dynamic Cheatsheet, Training-Free GRPO, and
EvoSkill fail because an unseen member is still counted. ReasoningBank fails just as often by dropping non-members
(Figure~\ref{fig:analysis}a in the appendix). Accuracy falls over the levels mostly because the share of single cells
grows (Figure~\ref{fig:retention} in the appendix). For ACE and Dynamic Cheatsheet, the exclusion of unseen rule-1
members at level 4 stays within 3 points of its level-1 value with both models. ACE with Luna moves from 15.0\% to
17.5\%, and with Sol from 72.5\% to 70.0\%. Keeping rule-1 non-members declines as rules accumulate, from 99.4\% to
77.5\% for Dynamic Cheatsheet with Sol and from 95.0\% to 84.4\% for ACE with Luna. Older rules lose specificity and
keep their members. Passing practice says little about learning. ACE passes 95\% of the practice levels and the history
reference 91\%, yet they solve 8\% and 13\% of the single cells, because the exception list is enough to pass.
Training-Free GRPO passes no practice level at all, so its mixed-outcome gate never opens. EvoSkill spends about 6M
input tokens per chain on practice and learns nothing. ACE spends about 0.9M, and the history reference spends 0.07M and
generalizes best among the self conditions.

\section{Conclusion}
\label{sec:conclusion}

\bench{} asks whether self-evolving agents learn from feedback a rule they were never told. Its hidden classes are
groupings whose members the base model knows, which it does not guess without feedback, and which it can identify from
the evidence once labels are read as named groupings. With the certifying model, current methods record corrections as exceptions and
gain at most 5 points over a frozen agent. A lesson that states the convention lifts them to 66--68\%, so the missing
step is inferring that corrections point to a named grouping. Gated updates starve under binary practice rewards, and
retrieved generic strategies over-generalize. Practice reward is a poor signal of learning: with the certifying model,
ACE passes 95\% of the practice levels yet solves 8\% of the single cells. A stronger base model, GPT-6-Sol, makes the
missing inference from feedback alone, and the benchmark can follow such progress as models improve. Three directions
follow from these results: update steps that propose named groupings and check them against the model's knowledge,
tools that let the agent look up published groupings during updates, and acceptance rules that do not depend on
reward variation. The construction pipeline lets others extend the benchmark to new domains and rerun it as methods
improve.

\bibliographystyle{ACM-Reference-Format}
\bibliography{refs}

\appendix
\section{Appendix}
\label{sec:appendix}

\begin{table*}[t]
  \caption{Hidden classes, one row per chain (levels 1--4). TR: Travel audit, SY: Pharmacy review, WS: Web security, LG: System logs, SP: Sponsorship, CH: Chemical imports, MN: Raw materials, SH: Port inspections, UN: Scholarships, GR: Travel grants.}
  \label{tab:classes}
  \scriptsize
  \setlength{\tabcolsep}{3pt}
  \begin{tabular}{@{}lp{0.225\textwidth}p{0.225\textwidth}p{0.225\textwidth}p{0.225\textwidth}@{}}
    \toprule
    Chain & Level 1 & Level 2 & Level 3 & Level 4 \\
    \midrule
    CH1 & Stockholm Convention Annex A & Commonwealth of Nations members & Star Alliance airlines & Chemical Weapons Convention Schedule 3 \\
    CH2 & Chemical Weapons Convention Schedule 3 & Group of Seven members & oneworld airlines & Stockholm Convention Annex A \\
    GR1 & Commonwealth of Nations members & Star Alliance airlines & League of European Research Universities members & NATO members \\
    GR2 & Schengen Area members & Five original Star Alliance airlines & ASEAN University Network members & ASEAN members \\
    LG1 & BRICS member country-code top-level domains & Office LTSC Professional Plus 2021 applications & Microsoft server product port assignments & Schengen Area member state country-code top-level domains \\
    LG2 & OECD member country country-code top-level domains & Microsoft productivity applications & Plaintext application protocol ports & G7 member country-code top-level domains \\
    MN1 & Commonwealth of Nations members & Star Alliance airlines & Yttrium-group rare-earth elements & G7 members \\
    MN2 & G7 members & SkyTeam airlines & Yttrium-group rare-earth elements & G20 members \\
    SH1 & G7 & NATO & Port locations recorded in Climate Alliance member cities & BRICS \\
    SH2 & Arab League & Asia-Pacific Economic Cooperation & Ports in countries belonging to BRICS & European Economic Area \\
    SP1 & European Cup and UEFA Champions League winners & EURO STOXX 50 constituents & Commonwealth of Nations members & 2025 FIFA Club World Cup participants \\
    SP2 & G-14 member clubs & Dow Jones Industrial Average constituents & European Union members & European Super League founding clubs \\
    SY1 & Inflammatory disorder of upper respiratory tract & Procedure on female genital system & Schedule II controlled substances & Infection \\
    SY2 & Degenerative disorder of musculoskeletal system & Procedure on uterus & Cytotoxic medicines & Inflammatory disorder of the respiratory tract \\
    TR1 & Founding airlines of Star Alliance & Airports in OPEC members & Currencies of BRICS members & Airlines of NATO members \\
    TR2 & Oneworld airlines & Airports in the European Economic Area & Currencies of Commonwealth of Nations members & Founding airlines of Oneworld \\
    UN1 & NATO members & ASEAN members & Nasdaq-100 constituents & G20 members \\
    UN2 & OECD members & Nasdaq-100 constituents & ASEAN University Network core members & G20 members \\
    WS1 & Comprehensive Categorization: Encryption (CWE-1402) & Italic languages & ASEAN members & Numeric Errors \\
    WS2 & SFP Primary Cluster: Entry Points & Official languages of the European Union & OPEC members & OWASP Top Ten 2025 Category A09:2025 - Logging \& Alerting Failures \\
    \bottomrule
  \end{tabular}
\end{table*}

\begin{figure*}[t]
  \centering
  \includegraphics[width=\textwidth]{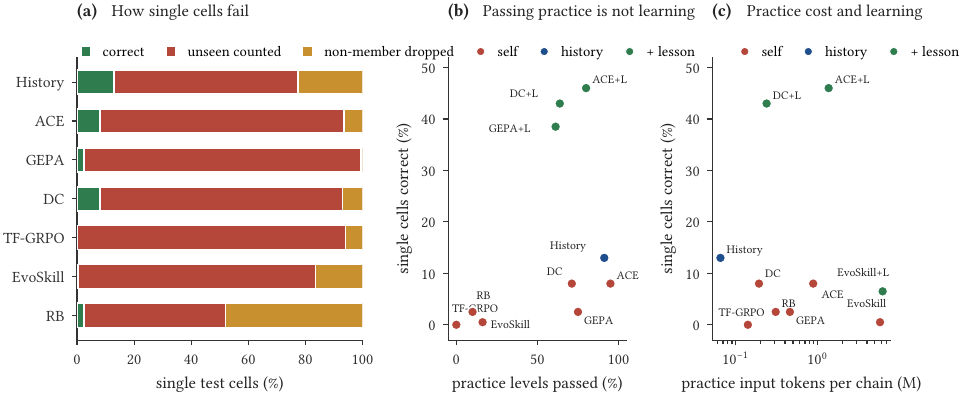}
  \caption{Failure modes and cost (GPT-6-Luna, learning from feedback alone unless marked +L). (a) How single cells fail. (b) Practice levels passed and (c) practice input tokens against single-cell accuracy.}
  \label{fig:analysis}
\end{figure*}

\begin{figure*}[t]
  \centering
  \includegraphics[width=\textwidth]{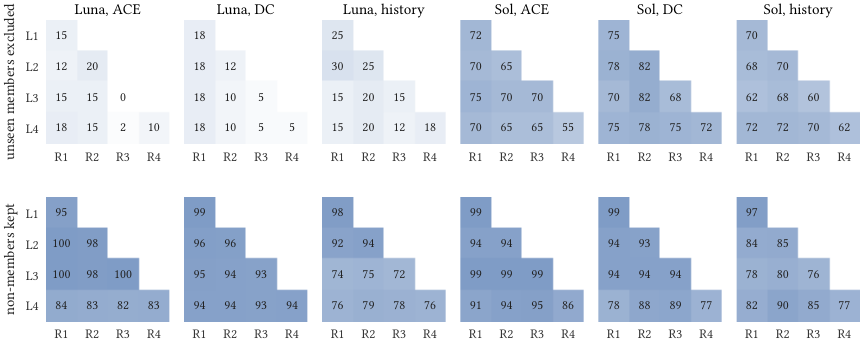}
  \caption{Retention: unseen members excluded (top) and non-members kept (bottom) for rule $j$ (columns) at level $k$ (rows).}
  \label{fig:retention}
\end{figure*}

\begin{table}[t]
  \caption{The ten domains: value-universe size and number of candidate lists per trap field.}
  \label{tab:domains}
  \small
  \begin{tabular}{@{}lp{0.62\columnwidth}@{}}
    \toprule
    Domain & Trap fields (values / lists) \\
    \midrule
    Travel audit & airline (1147/59), arrival airport (6675/27), currency (155/37) \\
    Pharmacy review & condition code (176/67), procedure code (165/56), ingredient (107/71) \\
    Web security & CWE (934/86), page language (177/63), host TLD (1285/40) \\
    System logs & TLD (1436/53), client application (131/36), port (6063/24) \\
    Sponsorship & club (151/12), co-sponsor (692/12), market country (249/10) \\
    Chemical imports & substance (654/18), exporter country (195/14), air carrier (239/7) \\
    Raw materials & origin country (194/16), carrier (452/14), material (118/19) \\
    Port inspections & flag state (250/32), owner country (250/32), port (751/26) \\
    Scholarships & nationality (249/22), referee country (249/22), sponsor (971/12) \\
    Travel grants & nationality (249/15), airline (1964/18), university (3762/15) \\
    \bottomrule
  \end{tabular}
\end{table}

\begin{table}[t]
  \caption{Conditions: text added to the update step (U) or the solver prompt (S).}
  \label{tab:conditions}
  \small
  \setlength{\tabcolsep}{3pt}
  \begin{tabular}{@{}lcll@{}}
    \toprule
    Condition & & Added & Tests \\
    \midrule
    Frozen & -- & nothing, no state & the frozen staircase \\
    Self (6 methods) & U & -- & induction from feedback \\
    \quad + lesson & U & convention + procedure & learnable from evidence \\
    \quad rewarded practice & -- & unseen $m_1$ from try 2 & response to failed lists \\
    History & S & raw practice history & update vs.\ raw evidence \\
    \quad + lesson & S & history + lesson & lesson, no method \\
    \midrule
    GPT-6-Sol & -- & stronger base model & model dependence \\
    \bottomrule
  \end{tabular}
\end{table}

\begin{table}[tp]
  \caption{All conditions, GPT-6-Luna (\%; $\Delta$: change over the frozen agent). Rows without + learn from feedback
  alone. $^\dagger$With a format note (Appendix~\ref{sec:gepa}). $^\ast$$m_1$ is
  revealed in practice; inc$_u$ counts $m_2$ only.}
  \label{tab:mainfull}
  \small
  \setlength{\tabcolsep}{2.6pt}
  \begin{tabular}{@{}lrrrrrrrrr@{}}
    \toprule
    & L1 & L2 & L3 & L4 & Avg & $\Delta$ & Single & inc$_u$ & exc \\
    \midrule
    Frozen & 80 & 57 & 34 & 20 & 47.8 & -- & 0 & 1 & 94 \\
    History & 84 & 65 & 37 & 23 & 52.3 & +4.5 & 13 & 20 & 82 \\
    \quad rewarded practice$^\ast$ & 90 & 75 & 51 & 39 & 63.8 & +16.0 & 31 & 42 & 87 \\
    History + lesson & 85 & 70 & 55 & 44 & 63.5 & +15.8 & 44 & 59 & 75 \\
    History + convention & 88 & 68 & 57 & 46 & 64.8 & +17.0 & 47 & 59 & 73 \\
    History + procedure & 82 & 67 & 40 & 28 & 54.3 & +6.5 & 18 & 32 & 82 \\
    Class-name oracle & 88 & 67 & 47 & 30 & 58.0 & +10.3 & 40 & 49 & 58 \\
    \midrule
    ACE & 79 & 64 & 44 & 23 & 52.5 & +4.8 & 8 & 12 & 92 \\
    \quad +\,lesson & 79 & 73 & 68 & 48 & 67.0 & +19.3 & 46 & 62 & 86 \\
    \quad +\,convention & 86 & 71 & 67 & 46 & 67.5 & +19.8 & 47 & 57 & 84 \\
    \quad rewarded practice$^\ast$ & 83 & 62 & 43 & 26 & 53.5 & +5.8 & 10 & 10 & 96 \\
    \addlinespace[2pt]
    GEPA$^\dagger$ & 81 & 58 & 37 & 22 & 49.5 & +1.8 & 3 & 6 & 96 \\
    \quad +\,lesson & 92 & 83 & 47 & 42 & 66.0 & +18.3 & 39 & 52 & 82 \\
    \quad +\,convention & 88 & 70 & 42 & 42 & 60.5 & +12.8 & 39 & 46 & 75 \\
    \quad rewarded practice$^\ast$ & 82 & 62 & 36 & 22 & 50.5 & +2.8 & 4 & 4 & 95 \\
    \addlinespace[2pt]
    Dynamic Cheatsheet & 83 & 62 & 42 & 24 & 52.8 & +5.0 & 8 & 12 & 95 \\
    \quad +\,lesson & 85 & 82 & 64 & 42 & 68.3 & +20.5 & 43 & 56 & 84 \\
    \quad +\,convention & 93 & 79 & 58 & 50 & 70.0 & +22.3 & 45 & 59 & 89 \\
    \quad rewarded practice$^\ast$ & 83 & 67 & 44 & 25 & 54.8 & +7.0 & 12 & 14 & 92 \\
    \addlinespace[2pt]
    Training-Free GRPO & 80 & 55 & 38 & 19 & 48.0 & +0.3 & 0 & 2 & 94 \\
    \quad +\,lesson & 80 & 60 & 39 & 19 & 49.5 & +1.8 & 1 & 1 & 96 \\
    \addlinespace[2pt]
    EvoSkill & 77 & 54 & 37 & 16 & 46.0 & -1.8 & 1 & 2 & 84 \\
    \quad +\,lesson & 80 & 48 & 26 & 14 & 42.0 & -5.8 & 7 & 10 & 62 \\
    \addlinespace[2pt]
    ReasoningBank & 68 & 36 & 15 & 15 & 33.5 & -14.3 & 3 & 9 & 51 \\
    \quad +\,lesson & 68 & 28 & 16 & 8 & 30.0 & -17.8 & 8 & 14 & 31 \\
    \bottomrule
  \end{tabular}
\end{table}

\begin{table}[t]
  \caption{Text added to the update step (GPT-6-Luna); columns as in Table~\ref{tab:mainfull}. Neutral heading:
  ``UPDATE NOTE'' instead of ``VERIFIED LESSON''.}
  \label{tab:variants}
  \small
  \setlength{\tabcolsep}{2.6pt}
  \begin{tabular}{@{}llrrrrrrrr@{}}
    \toprule
    & Update-step text & L1 & L2 & L3 & L4 & Avg & Single & inc$_u$ & exc \\
    \midrule
    ACE & procedure, neutral heading & 83 & 65 & 45 & 27 & 55.0 & 10 & 15 & 98 \\
     & convention, neutral heading & 91 & 85 & 65 & 45 & 71.5 & 47 & 57 & 91 \\
     & lesson, neutral heading & 89 & 83 & 66 & 50 & 72.0 & 48 & 60 & 94 \\
     & convention at level 1 only & 83 & 75 & 58 & 39 & 63.8 & 34 & 47 & 94 \\
     & class names + feedback & 93 & 82 & 64 & 55 & 73.5 & 52 & 66 & 88 \\
    \addlinespace[2pt]
    DC & procedure, neutral heading & 77 & 64 & 46 & 33 & 55.0 & 18 & 28 & 90 \\
     & convention, neutral heading & 87 & 73 & 61 & 42 & 65.8 & 38 & 50 & 87 \\
     & lesson, neutral heading & 82 & 63 & 44 & 26 & 53.8 & 15 & 20 & 81 \\
     & convention at level 1 only & 88 & 72 & 56 & 41 & 64.3 & 33 & 38 & 88 \\
     & class names + feedback & 94 & 83 & 68 & 58 & 75.8 & 57 & 73 & 87 \\
    \bottomrule
  \end{tabular}
\end{table}

\begin{table}[t]
  \caption{Other base models (\%). Frozen: accuracy per level and level scores above the staircase (of 80). Oracle:
  the class-name oracle.
  $^\ast$Evaluated and certifying model.}
  \label{tab:backbones}
  \small
  \setlength{\tabcolsep}{2.4pt}
  \begin{tabular}{@{}lccrrrr@{}}
    \toprule
    & Frozen & $>$stair & Oracle & Single & inc$_u$ & exc \\
    \midrule
    GPT-6-Luna (high)$^\ast$ & .80/.57/.34/.20 & 0 & 58.0 & 40 & 49 & 58 \\
    GPT-5.6-Luna (high) & .80/.60/.40/.18 & 0 & 62.8 & 45 & 54 & 63 \\
    GPT-5.6-Terra (high) & .76/.48/.38/.14 & 1 & 57.0 & 42 & 49 & 51 \\
    Claude Sonnet 5.5 (low) & .80/.55/.27/.20 & 4 & -- & -- & -- & -- \\
    GPT-6-Sol (high) & .77/.32/.15/.06 & 1 & -- & -- & -- & -- \\
    \bottomrule
  \end{tabular}
\end{table}

\begin{table}[t]
  \caption{Both base models (\%). Methods learn from feedback alone unless marked + lesson. $^\dagger$With a format note (Appendix~\ref{sec:gepa}).}
  \label{tab:sol}
  \small
  \setlength{\tabcolsep}{2.6pt}
  \begin{tabular}{@{}lrrrrrrrr@{}}
    \toprule
    & \multicolumn{4}{c}{GPT-6-Luna} & \multicolumn{4}{c}{GPT-6-Sol} \\
    \cmidrule(lr){2-5}\cmidrule(l){6-9}
    & Acc & Single & inc$_u$ & exc & Acc & Single & inc$_u$ & exc \\
    \midrule
    Frozen & 47.8 & 0 & 1 & 94 & 32.5 & 1 & 6 & 39 \\
    \addlinespace[2pt]
    History & 52.3 & 13 & 20 & 82 & 70.3 & 49 & 68 & 83 \\
    \quad + lesson & 63.5 & 44 & 59 & 75 & 83.8 & 71 & 84 & 91 \\
    \addlinespace[2pt]
    ACE & 52.5 & 8 & 12 & 92 & 78.3 & 59 & 68 & 95 \\
    \quad + lesson & 67.0 & 46 & 62 & 86 & 81.3 & 68 & 82 & 92 \\
    Dynamic Cheatsheet & 52.8 & 8 & 12 & 95 & 78.5 & 60 & 76 & 90 \\
    \quad + lesson & 68.3 & 43 & 56 & 84 & 84.0 & 70 & 83 & 95 \\
    \addlinespace[2pt]
    GEPA$^\dagger$ & 49.5 & 3 & 6 & 96 & 58.8 & 37 & 48 & 66 \\
    \quad + lesson & 66.0 & 39 & 52 & 82 & 64.8 & 48 & 60 & 67 \\
    Training-Free GRPO & 48.0 & 0 & 2 & 94 & 35.8 & 2 & 6 & 46 \\
    \quad + lesson & 49.5 & 1 & 1 & 96 & 31.2 & 1 & 4 & 36 \\
    EvoSkill & 46.0 & 1 & 2 & 84 & 52.3 & 35 & 41 & 53 \\
    \quad + lesson & 42.0 & 7 & 10 & 62 & 70.5 & 52 & 65 & 76 \\
    ReasoningBank & 33.5 & 3 & 9 & 51 & 24.8 & 4 & 7 & 21 \\
    \quad + lesson & 30.0 & 8 & 14 & 31 & 20.8 & 7 & 11 & 14 \\
    \bottomrule
  \end{tabular}
\end{table}

\begin{table}[t]
  \caption{Average test accuracy (\%, levels 1--4) per domain, two chains each. +L: with the lesson. The lesson adds little in web security and pharmacy review.}
  \label{tab:domainres}
  \small
  \setlength{\tabcolsep}{3pt}
  \begin{tabular}{@{}lrrrrrrr@{}}
    \toprule
    Domain & Frozen & History & ACE & +L & DC & +L & Hist.+L \\
    \midrule
    Travel audit & 50.0 & 57.5 & 70.0 & 70.0 & 50.0 & 72.5 & 87.5 \\
    Pharmacy review & 50.0 & 35.0 & 47.5 & 55.0 & 50.0 & 55.0 & 35.0 \\
    Web security & 50.0 & 52.5 & 50.0 & 40.0 & 50.0 & 47.5 & 70.0 \\
    System logs & 45.0 & 50.0 & 47.5 & 67.5 & 50.0 & 67.5 & 85.0 \\
    Sponsorship & 32.5 & 47.5 & 47.5 & 65.0 & 52.5 & 65.0 & 70.0 \\
    Chemical imports & 50.0 & 55.0 & 50.0 & 62.5 & 62.5 & 67.5 & 40.0 \\
    Raw materials & 50.0 & 70.0 & 50.0 & 80.0 & 62.5 & 85.0 & 75.0 \\
    Port inspections & 50.0 & 50.0 & 50.0 & 77.5 & 50.0 & 90.0 & 82.5 \\
    Scholarships & 50.0 & 47.5 & 50.0 & 75.0 & 50.0 & 65.0 & 25.0 \\
    Travel grants & 50.0 & 57.5 & 62.5 & 77.5 & 50.0 & 67.5 & 65.0 \\
    \bottomrule
  \end{tabular}
\end{table}

\subsection{State annotation}
\label{sec:annotation}
For every chain and level we take the state a method holds after practicing level $k$ and annotate what it says about
the label of level $k$. The annotator (GPT-6-Luna, high reasoning effort) sees the field, the label, the hidden class
and its members, and the values revealed in practice, but no test program or score. It assigns one of seven categories:
the exact class or an equivalent description; a broader grouping that contains the class; another named grouping; an
attribute rule (region, country, size, name pattern); only the values seen in practice; most members of the class
listed without naming it; or nothing usable. We annotate 1,280 states: ACE and Dynamic Cheatsheet learning from
feedback alone, with the lesson, and under rewarded practice with GPT-6-Luna; with the procedure, the convention, and
the whole lesson under the neutral heading with GPT-6-Luna; and from feedback alone and with the lesson with GPT-6-Sol.
Each state is then linked to the value-level diagnostics of the program written from it at the same level.

\subsection{The lesson}
The +\,lesson condition appends the following text, under the heading ``VERIFIED LESSON (correct, from the task
designers)'', to the input of each method's update step; the History + lesson reference places it in the
solver prompt. The text is the same for every chain and level and does not name any class. It states a convention (labels
denote well-known named groupings rather than attribute-defined categories) and a procedure for using the evidence.

\begin{promptbox}{Lesson}
When a level rule applies to a labelled class and the report gives only one example value, the class is a real-world
category that the label does not name: a named grouping that exists in this domain, such as an organisation, an
alliance, an official list or a standard classification. Feedback that another value is treated like the example, and
that some other value is not, tells you which category it is: choose the best-known named grouping that contains the
example and the confirmed value and excludes the rejected one. Do not construct a category from a property you read in
the descriptions, such as a region, a type of government, a climate or a size. Write that category by name into your
notes together with the label. Then apply the rule to every value the report describes that belongs to that grouping,
using each description only to identify what the value is and your knowledge of the grouping to decide membership,
including values that never appeared in feedback, and leave out values that are not in the grouping. Before coding, go
through every described value, decide membership for each one, and hard-code the complete member set.
\end{promptbox}

\paragraph{Lesson parts.} History + convention uses the first three sentences of the lesson (up to ``a climate or a
size''). History + procedure uses: ``When a level rule applies to a labelled class and the report gives only one example
value, decide which class the example and the feedback point to and write it into your notes together with the label.
Then apply the rule to every value the report describes that belongs to that class, including values that never
appeared in feedback, and leave out values that are not in it. Before coding, go through every described value, decide
membership for each one, and hard-code the complete member set.''

\subsection{Implementation details}
All methods use the official prompts and algorithms with task-specific placeholders filled in; the frozen agent uses the
same solver prompt with an empty state. A practice attempt is one submitted program. Training-Free GRPO samples a group of four rollouts within each
attempt and, as in the official
implementation, extracts experiences only from groups with mixed outcomes. ReasoningBank retrieves the top memory items
by embedding similarity with a local sentence-transformers model. EvoSkill's skill library can exceed the context window
at the last levels; in that case we fall back to the previous level's state and record the fallback. In the +\,lesson
condition, the lesson is appended to every update input and nothing else changes. The history reference shows the solver every earlier practice program, from all levels so far,
with its feedback; the class-name oracle adds one line per label (``review queue B8 stands for: founding member
airlines of Star Alliance'') and an instruction to apply each rule to exactly the members of the named grouping.

\paragraph{GEPA note.}\label{sec:gepa}
GEPA folds every detail of the practice task into its evolved prompt, including the practice aggregation (largest
total), which then overrides the test task (sum). All GEPA results in the main text therefore append one fixed note to
its reflection input: ``Note on the policy: it will also be used for this level's test task, which has the same records
and rules but may ask for a different output (for example a different aggregation over the entities). Keep the policy
about which records count and which rules apply; do not state the output aggregation or format, and let each task's
own instructions decide them.'' Under explicit feedback, the original GEPA without any note reaches \gepaorigavg\% average accuracy (self), with
\gepaorigmax{} of its 160 test programs (self and +\,lesson) returning the practice aggregation. A variant whose note gives the test
aggregation as its example (``for example the sum over all entities instead of the largest total'') reaches 50.8\%
(self) and 62.5\% (+\,lesson); the neutral note of the main results reaches \gepaneutralselfavg\% and
\gepaneutralhintavg\%.

\subsection{Example task and feedback}
\label{sec:example}
Excerpts from the level-1 practice task of chain TR1 and the feedback on a first attempt that excluded no airline.

\begin{promptbox}{Practice task (excerpt)}
Each flight segment on a submitted expense claim counts once toward the total of the department of the employee who
filed that claim. Return the largest total among the departments in the input, as an int.\par\smallskip
Field notes:\\
- A segment whose airline is 'Lufthansa' is in review queue B8.\\
Each label in the field notes stands for a real-world category of values; a note gives one member of that category,
and the label's name does not say which category it is.\par\smallskip
Level rules:\\
L1: From this level on, segments in review queue B8 are not counted.\par\smallskip
Value reference:\\
- Air Canada -- Airline based in Canada.\\
- ITA Airways -- Airline based in Italy.\\
- Thai Airways -- Airline based in Thailand.\\
- United Airlines -- Airline based in US.\\
\dots
\end{promptbox}

\begin{promptbox}{Feedback on attempt 1 (excerpt)}
Wrong. Your submission gives a wrong result on the input below, which is one of the hidden test inputs.
\dots\par\smallskip
Compared with the example input, this input differs only in: segments.airline: one value 'Air Dabia' was replaced by
'Thai Airways'; segments.airline: one value 'Air Burkina' was replaced by 'ITA Airways'.\\
'Thai Airways' belongs to review queue B8, like 'Lufthansa', so the record with this value is not counted.\\
'ITA Airways' does not belong to review queue B8, so the record with this value is still counted.
\end{promptbox}

\end{document}